\documentclass[sigconf]{acmart}

\AtBeginDocument{%
  \providecommand\BibTeX{{%
    \normalfont B\kern-0.5em{\scshape i\kern-0.25em b}\kern-0.8em\TeX}}}
 
\copyrightyear{2024}
\acmYear{2024}
\setcopyright{rightsretained}
\acmConference[KDD '24]{Proceedings of the 30th ACM SIGKDD Conference on Knowledge Discovery and Data Mining}{August 25--29, 2024}{Barcelona, Spain}
\acmBooktitle{Proceedings of the 30th ACM SIGKDD Conference on Knowledge Discovery and Data Mining (KDD '24), August 25--29, 2024, Barcelona, Spain}\acmDOI{10.1145/3637528.3671527}
\acmISBN{979-8-4007-0490-1/24/08}

\makeatletter
\gdef\@copyrightpermission{
 \begin{minipage}{0.3\columnwidth}
  \href{https://creativecommons.org/licenses/by/4.0/}{\includegraphics[width=0.90\textwidth]{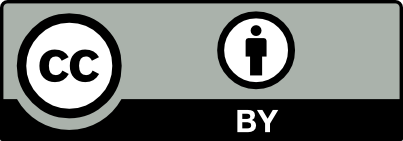}}
 \end{minipage}\hfill
 \begin{minipage}{0.7\columnwidth}
  \href{https://creativecommons.org/licenses/by/4.0/}{This work is licensed under a Creative Commons Attribution International 4.0 License.}
 \end{minipage}
 \vspace{5pt}
}
\makeatother

\newcommand{\mysystem}{VecAug}
\DeclareMathOperator{\dis}{d}
\usepackage{multirow}
\usepackage{makecell}
\usepackage{enumitem}
\usepackage{color}
\usepackage{graphicx}
\usepackage[super]{nth}
\usepackage{bm}
\theoremstyle{definition}
\newtheorem{definition}{Definition}
\newcolumntype{P}[1]{>{\centering\arraybackslash}p{#1}}
\usepackage[linesnumbered,ruled,noend]{algorithm2e}
\usepackage{comment}
\usepackage[a-2b]{pdfx}

\newcommand{\highlight}[1]{\textbf{#1}}

\usepackage{amsbsy}

\usepackage{balance}

\begin{document}

\title{{\mysystem}: Unveiling Camouflaged Frauds with Cohort Augmentation for Enhanced Detection}

\author{Fei Xiao}
\orcid{0009-0003-1493-8040}
\affiliation{%
  \institution{National University of Singapore}
  \institution{\& Shopee Singapore}
}
\email{fxiao004@comp.nus.edu.sg}

\author{Shaofeng Cai}
\authornote{Corresponding author}
\affiliation{%
  \institution{National University of Singapore} 
  \country{Singapore}
}
\email{shaofeng@comp.nus.edu.sg}

\author{Gang Chen}
\affiliation{%
  \institution{Zhejiang University}
  \country{Hangzhou, China}
}
\email{cg@zju.edu.cn}

\author{H. V. Jagadish}
\affiliation{%
  \institution{University of Michigan}
  \country{Ann Arbor, USA}
}
\email{jag@umich.edu}

\author{Beng Chin Ooi}
\affiliation{%
  \institution{National University of Singapore}
  \country{Singapore}
}
\email{ooibc@comp.nus.edu.sg}

\author{Meihui Zhang}
\affiliation{%
  \institution{Beijing Institute of Technology}
  \country{Beijing, China}
}
\email{meihui_zhang@bit.edu.cn}



\begin{abstract}
Fraud detection presents a challenging task characterized by ever-evolving fraud patterns and scarce labeled data.
Existing methods predominantly rely on graph-based or sequence-based approaches.
While graph-based approaches connect users through shared entities to capture structural information, they remain vulnerable to fraudsters who can disrupt or manipulate these connections.
In contrast, sequence-based approaches analyze users' behavioral patterns, offering robustness against tampering but overlooking the interactions between similar users.
Inspired by cohort analysis in retention and healthcare, this paper introduces {\mysystem}, a novel cohort-augmented learning framework that addresses these challenges by enhancing the representation learning of target users with personalized cohort information.
To this end, we first propose a \textit{vector burn-in} technique for automatic \textit{cohort identification}, which retrieves a task-specific cohort for each target user.
Then, to fully exploit the cohort information, we introduce an \textit{attentive cohort aggregation} technique for augmenting target user representations.
To improve the robustness of such cohort augmentation, we also propose a novel \textit{label-aware cohort neighbor separation} mechanism to distance negative cohort neighbors and calibrate the aggregated cohort information.
By integrating this cohort information with target user representations, {\mysystem} enhances the modeling capacity and generalization capabilities of the model to be augmented.
Our framework is flexible and can be seamlessly integrated with existing fraud detection models.
We deploy our framework on e-commerce platforms and evaluate it on three fraud detection datasets, and results show that {\mysystem} improves the detection performance of base models by up to 2.48\% in AUC and 22.5\% in R@P$_{0.9}$, outperforming state-of-the-art methods significantly.

\end{abstract}



\maketitle

\section{Introduction}


Fraudulent transactions pose a serious threat to e-commerce platforms, as they cause economic losses, damage reputations, and harm user experiences~\cite{Liu2020, Wang2023}.
According to a recent report, fraud costs 2.9\% of global e-commerce revenue in 2023~\footnote{\url{https://www.identiq.com/blog/ecommerce-fraud-prevention-report}}.
This is especially concerning for the fast-growing Southeast Asian market, which is expected to have 140 million new consumers by 2030~\footnote{\url{https://www.linkedin.com/pulse/why-southeast-asia-expected-worlds-next-e-commerce-powerhouse/}}. 
However, detecting fraud in real-world scenarios remains challenging due to the dynamic nature of fraud patterns and the scarcity of labeled data~\cite{Wang2023}.

To address the sparse transaction data issue and identify potential fraudsters among new users, numerous fraud detection systems have been developed, leveraging advanced machine learning algorithms and integrating diverse data sources for comprehensive analysis and detection. These systems typically fall into two categories: graph-based approaches~\cite{Liu2018, Wang2019b, Cheng2020a, Dou2020, Liu2021b, Zhang2023b, Huang2022, Shi2022} and sequence-based methodologies~\cite{Wang2017, Li2018, Zhu2020, Liu2020, Branco2020, Lin2021, Liu2021, Wang2023}. 
Graph-based systems connect users based on shared identity features, such as devices, IPs, and phones, and use graph neural networks (GNNs) to capture the structural information. 
However, certain fraudsters employ sophisticated techniques, such as device information manipulation or VPN usage, to evade explicit connections.
They also imitate the behaviors of legitimate users, such as clicking on the same products or pages or using similar IPs, to camouflage themselves~\cite{Liu2018, Dou2020}.
These fraudsters are hard to identify using graph-based methods and lower the effectiveness of GNN-based approaches. 
In contrast, sequence-based systems analyze users’ behavioral sequences, which are more resilient to manipulation than individual features~\cite{Liu2020, Liu2021, Xiao2023}.
Nonetheless, sequence-based models often overlook cross-user interactions between the target user and similar peers, thus failing to fully exploit the information from similar users.
Therefore, we need tailored approaches to connect well-camouflaged fraudsters with similar users and use their data for more accurate user representations.

\begin{figure} 
\centering
\includegraphics[scale=0.23]{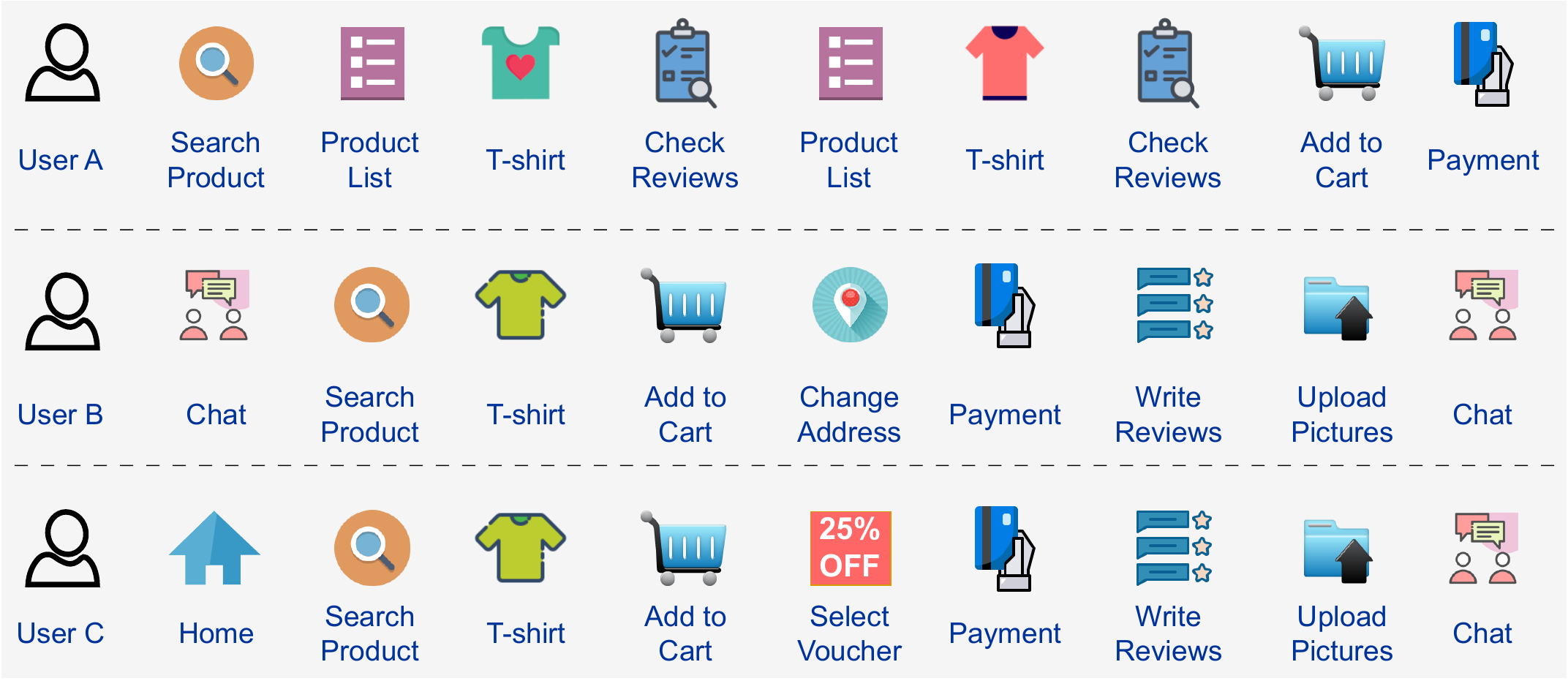}
\caption{A toy example depicting the behavioral sequences of three users: User B and User C have similar behavior sequences, whereas User A has quite different ones.
}
\label{fig: example}
\end{figure}

In this paper, we present {\mysystem}, a novel learning-based cohort augmentation framework for fraud detection using cohort information.
Our framework is inspired by cohort analysis, which is widely used in retention and healthcare applications to study the relationships and behaviors of users based on their activity patterns.
Unlike traditional cohort analysis methods that group users into different cohorts based on a set of traits or a specific event, our framework constructs implicit user cohorts based on the relevance of their behavioral sequences.
In this manner, we can identify potential fraudsters who exhibit similar fraudulent behaviors, regardless of their identity features, such as devices, IPs, and phones.
For example, Figure~\ref{fig: example} illustrates the behavioral sequences of three users, where User B and User C both search for the same products, make quick payments, and leave positive reviews.
They can be regarded as personalized cohort neighbors of each other and their information can be shared to enhance the homophily for fraud prediction. 
In contrast, User A performs product searches and comparisons before buying, behaving more like a normal user and should not be included in their cohort neighbors.


As illustrated in Figure~\ref{fig: vecaug}, {\mysystem} has four phases: the vector burn-in phase, the cohort identification phase, the cohort-augmented training phase, and the prediction phase. 
In the vector burn-in phase, we use a novel technique to generate task-specific cohort vectors for all training samples and store them in a vector database $\mathcal{E}_{pool}$ along with their fraud labels. 
This technique separates the optimization of embedding vectors from their use in the later cohort augmentation, which improves the training efficiency and effectiveness. 
In the cohort identification phase, we select $K$ augmentation neighbors and $K$ negative neighbors for each target user from the vector pool, based on their similarity and label discrepancy. 
The augmentation neighbors are used to enrich the representation of the target user, while the negative neighbors are used to create a contrastive learning objective. 
In the cohort-augmented training phase, {\mysystem} performs attentive neighbor aggregation to combine the cohort information from the augmentation neighbors, and label-aware neighbor separation to push away the representations of the negative neighbors. 
By doing so, {\mysystem} learns a more robust and discriminative model for fraud detection. In the prediction phase, {\mysystem} can rapidly retrieve the augmentation neighbors from the vector pool and use them to improve the prediction of the target user.
In summary, we make the following contributions:
\begin{itemize}[leftmargin=*] 
    \item 
    We introduce {\mysystem}, a pioneering cohort augmentation framework that improves the prediction performance of existing fraud detection methods by incorporating rich cohort neighborhood information and task-specific embedding vectors.
    \item
    We develop a novel vector burn-in technique that identifies personalized cohorts based on behavioral sequences and detects fraudsters who evade connections by identity features.
    \item
    We propose a novel label-aware neighbor separation technique that reduces noise from neighbors of opposite classes and improves cohort-augmented learning robustness.
    \item 
    We test {\mysystem} on the state-of-the-art deep learning models in fraud detection and observe substantial performance improvement. {\mysystem} enhances base models by up to 2.48\% in AUC and 22.5\% in R@P$_{0.9}$, with negligible computational overhead.
    
\end{itemize}

We organize this paper as follows. Section \ref{sec: problem} introduces the fraud detection tasks and the cohort augmentation concept. Section \ref{sec: method} explains {\mysystem}'s technical aspects and shows how personalized cohorts help detect camouflaged fraudsters. 
Section \ref{sec: experiment} presents experimental results of integrating {\mysystem} with four base models to demonstrate its effectiveness. Section \ref{sec: related} discusses the related work on fraud detection, and section \ref{sec: conclusion} concludes this paper.

\section{Problem Definition}
\label{sec: problem}

In this section, we first formulate the fraud detection task on user behavioral data.
Next, we introduce the idea of learning-based cohort augmentation for fraud detection, followed by the definition of \textit{augmentation neighbors} and \textit{negative neighbors}. 

\subsection{Fraud Detection}
\label{subsec: problem_statement}

\noindent
\highlight{Fraud Detection on Behavioral Data.}
For a target user $u_i \in \mathcal{U}$, a common fraud behavior detection system takes the user's behavioral sequence as input: $\mathbf{x_i} = \{s_0, s_1, \cdots, s_{m-1} \}$, where $s_i \in \mathcal{S}$ represents user's action and $m$ is the length of the sequence.
The fraud detection task over behavioral data aims to detect whether users engage in suspicious activities given their behavioral sequences.
Each user $u_i$ is characterized by a corresponding behavioral sequence $\mathbf{x_i}$ and a ground-truth label $y_i$.
Identifying fake reviews is a specific instance of fraud behavior detection, where actions correspond to purchases of different products.

\begin{figure} 
\centering
\includegraphics[scale=1.2]{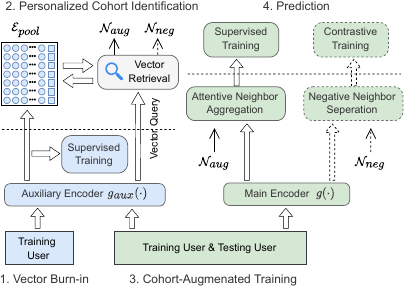}
\caption{The proposed learning-based cohort augmentation framework {\mysystem}. {\mysystem} seamlessly integrates with any existing fraud detection models to uncover hidden connections between users and utilize cohort information for enhanced fraud detection. 
}
\label{fig: vecaug}
\end{figure}

\subsection{Cohort Augmentation for Fraud Detection}
In the cohort augmentation framework {\mysystem}, we retrieve a personalized cohort for each user, aiming to exploit the implicit relationships between target users and their hidden connections.
Specifically, the personalized cohort comprises $K$ \textit{augmentation neighbors}, which are utilized to enhance target user representations, and $K$ \textit{negative neighbors}, which contribute to the creation of an effective contrastive loss for distinguishing neighbors with different labels from the target user.

\begin{definition}\textbf{Learning-based Cohort Augmentation.}
Given the behavioral sequence of $u_i$ in the e-commerce platform as $\mathbf{x}_i = \{s_1, s_2, \cdots, s_{n} \}$ and the corresponding label $\mathbf{y}_i$,
the objective of cohort augmentation is to pinpoint similar fraudulent users controlled by the same entity as the target fraud users, and to identify additional normal users who resemble the target normal user, leveraging their information to enhance the predictive performance of the model.
We can formulate learning-based cohort augmentation as:

\begin{equation}
      \widehat{\mathbf{y}}_i^{a} = f(g(\mathbf{x}_{i}; \mathbf{W}_g); \mathbf{e}_i; \textbf{E}_i^{a}; \mathbf{W}_f),
\label{eq:f_LC_x}
\end{equation}

\noindent where $f(\cdot)$ is the augmented learning function parameterized by $\mathbf{W}_f$; $g(\cdot)$ is the representation learning function of the base model with parameters $\mathbf{W}_g$;
$\mathbf{e}_i$ and $\textbf{E}_i^{a}$ are the vector of the target user and the set of task-specific cohort augmentation neighbors in the vector burn-in space respectively, derived by a separately trained vector burn-in model;
$\widehat{\mathbf{y}}_i^{a}$ represents the cohort-augmented prediction logit.
The main motivation for introducing the personalized learning-based cohorts is to leverage the supplementary information of similar samples $\textbf{E}_i^{a}$ for enhancing the representation power and robustness of the representation learning function $g(\cdot)$.
\end{definition}

\begin{definition}\textbf{Augmentation Neighbors.}
The augmentation neighbors, denoted as $\mathcal{N}_{aug}(u_i, K)$, are $K$ training samples that have similar fraud behaviors to the target sample $u_i$, as measured by the distance between their burn-in vectors. 
The augmentation neighbors can have the same or different labels as $u_i$ (fraudulent or non-fraudulent). The augmentation neighbors help to improve the representation learning of the target sample by providing cohort information. The set of burn-in vectors of the augmentation neighbors is denoted as $\textbf{E}_i^{a}$.
\end{definition}

\begin{definition}\textbf{Negative Neighbors.}
The negative neighbors, denoted as $\mathcal{N}_{neg}(u_i, K)$, are $K$ training samples that have opposite labels from the target sample $u_i$ but have similar fraud behaviors, based on the distance between their burn-in vectors. The negative neighbors are used to improve the robustness and stability of cohort-augmented learning by minimizing the mutual information between the cohort data and the target user. The set of burn-in vectors of the negative neighbors is denoted as $\textbf{E}_i^{n}$.
\end{definition}
\section{Methodologies}
\label{sec: method}
This section introduces {\mysystem}, a novel framework that aims to boost the performance of existing fraud detection methods by enhancing their discrimination capability. 
We describe each module of {\mysystem} and how they are trained to achieve this goal.

\begin{figure} 
\centering
\includegraphics[scale=1.2]{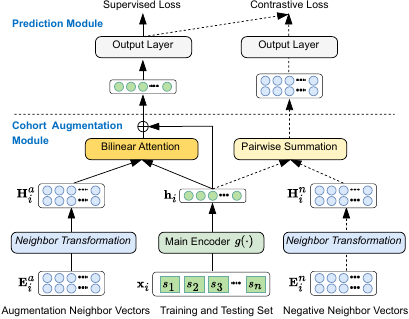}
\caption{The detailed illustration of cohort augmentation module and prediction module. The neighbor transformation blocks are identical. The dashed arrows denote computational flows involved only in the augmented training phase.}
\label{fig: CoNN}
\end{figure}

\subsection{Overall Framework}
The illustration of {\mysystem} for enhanced fraud detection tasks in Figure \ref{fig: vecaug} presents the four main modules operating in distinct phases:
the vector burn-in phase, the cohort identification phase, the augmented training phase, and the prediction phase.
First, the vector burn-in module will generate fraud-specific vectors $\mathbf{e}_i$ via an auxiliary encoder $g_{aux}(\cdot)$ and store the resultant embedding vectors of all training samples in the vector database $\mathcal{E}_{pool}$.
For each user $u_i$, {\mysystem} identifies its $K$ augmentation neighbor vectors $\textbf{E}_i^{a}$ and $K$ negative neighbor vectors $\textbf{E}_i^{n}$ from $\mathcal{E}_{pool}$.
These vectors capture the relationship between users based on their fraud behaviors.
Next, in the augmented training phase as shown in Figure~\ref{fig: CoNN}, {\mysystem} aggregates the information of $\textbf{E}_i^{a}$ to augment the latent representation $\mathbf{h}_i$ of $\mathbf{x}_i$ obtained via the main encoder $g(\cdot)$.
To address camouflaged frauds, {\mysystem} applies a label-aware neighbor separation mechanism to push away the $K$ negative neighbors in the augmented representation space, which reduces the influence of fraudsters who imitate normal users and improves detection accuracy.
Finally, in the prediction phase, only the augmentation neighbor vectors $\textbf{E}_i^{a}$ will be retrieved to augment the target user representation $\mathbf{h}_i$ for the final prediction.
For clarity, the outputs of the auxiliary encoder and the main encoder are called \textit{vector} and \textit{representation}, respectively.

\subsection{Vector Burn-in Phase}
\label{subsec: Burn-in}
To generate embedding vectors that can identify cohorts of fraudsters, we propose a novel vector burn-in technique that captures the fraud behaviors within each sample.
Specifically, we train an auxiliary encoder $g_{aux}(\cdot)$ and an output layer $\phi_{aux}$ using back-propagation in a supervised manner to obtain embedding vectors, i.e., $\mathbf{e}_i = g_{aux}(\mathbf{x}_i)$.
These vectors encode fraud-related information that is useful for subsequent cohort-augmented learning.
The auxiliary encoder can leverage existing fraud detection models, such as those proposed in~\cite{Zhu2020, Liu2020, Xiao2023} or traditional sequence learning methods in~\cite{Schuster1997, Velickovic2018}, to obtain the fraud-specific embedding of samples and use a multilayer perceptron (MLP) as the output layer to generate prediction logits:

\begin{equation}
    \widehat{\mathbf{y}}_i = \phi_{aux}(\mathbf{e}_i), \quad \phi_{aux}:\mathbb{R}^{n_d} \mapsto \mathbb{R}
\end{equation}

\noindent where $n_d$ is the dimension of the embedding vectors. 
Binary cross entropy (BCE) is typically adopted as the objective function for the fraud detection task and is optimized via gradient descent.
In this way, task-specific embedding vectors can be generated, and the corresponding latent embedding space is called \textit{vector burn-in space}, which will not be optimized in the augmented training phase.
The novel vector burn-in technique effectively decouples cohort identification from the base model training process, eliminating early training noise, and ensuring more accurate cohort neighbors.

\begin{figure} 
\centering
\includegraphics[scale=0.8]{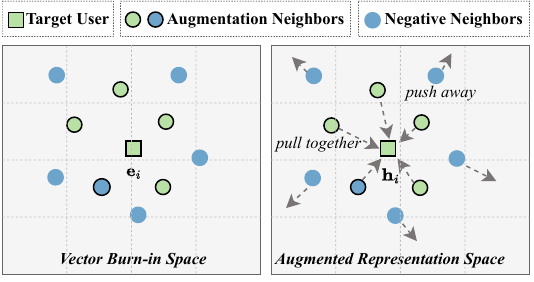}
\caption{The augmentation neighbors and negative neighbors for the target sample $u_i$ in vector burn-in space and augmented representation space.}
\label{fig: label-aware}
\end{figure}

\subsection{Cohort Identification Phase}
The embedding vectors of training samples, denoted as $\mathbf{e}_i$, along with their corresponding labels $\mathbf{y}_i$, are stored in a vector database $\mathcal{E}_{pool}$.
To identify the cohorts specific to fraudulent activities, the target sample is fed into $g_{aux}(\cdot)$, generating a vector query to search for $K$ augmentation neighbors in $\mathcal{E}_{pool}$.
During this phase, the Euclidean distance metric is employed for retrieving the cohort neighbors.
However, cohort augmentation for fraud detection faces a key challenge: the potential degradation of prediction performance.
This can occur when incorporating information from well-camouflaged fraudsters with that of normal users, or when excessively integrating normal user behavior information to enhance the representation learning of fraudsters. 
To mitigate this challenge, we adopt a strategy that involves retrieving both $K$ augmentation neighbors and $K$ negative neighbors for each target user during training.
This strategy enables the model to discern the behavior disparities between well-camouflaged fraudsters and normal users.
We construct a tailored contrastive loss to achieve this goal, which is illustrated in Figure~\ref{fig: CoNN} and explained in Section~\ref{sec:multi_objective_training}. 
Here, $\textbf{E}_{i}^{a}$ and $\textbf{E}_{i}^{n}$ represent the augmentation neighbor vectors and negative neighbor vectors, respectively.
It is worth noting that during inference for augmented learning, only the augmentation neighbors $\textbf{E}_{i}^{a}$ are retrieved, as the labels of testing samples are not available to identify the negative neighbors $\textbf{E}_{i}^{n}$.

\subsection{Cohort-Augmented Training Phase}
To harness the potential of neighboring vector information and effectively augment the target sample modeling, we perform an \textit{attentive neighbor aggregation} for the $K$ augmentation neighbors and propose a label-aware neighbor separation mechanism to distance the $K$ negative neighbors.
We first transform the augmentation neighbor vectors and the negative neighbor vectors to the \textit{augmented representation space}, which is the new representation space derived from the main encoder $g(\cdot)$:

\begin{equation}
    \textbf{H}_{i}^{a}= \mathbf{W}^{aug} \textbf{E}_{i}^{a} + \mathbf{b}^{aug}
\label{eq:trasnformation}
\end{equation}

\noindent where $\mathbf{W}^{aug} \in \mathbb{R}^{n_d \times n_d}$ is the weight vector, $\mathbf{b}^{aug} \in \mathbb{R}^{n_d}$ is the bias term, and $\textbf{H}_{i}^{a}$ is the representations of the augmentation neighbors.
Then we attentively aggregate information from the augmentation neighbors to the target sample using the following equations:

\begin{equation}
\begin{gathered}
    \alpha_{i,k} = \text{Softmax}(\mathbf{h}_{i,k}^{T}\textbf{W}^{att} \mathbf{h}_{i}), \quad \mathbf{h}_{i,k} \in \textbf{H}_{i}^{a} \\
    \mathbf{h}_{i}^{a} = \sum_{k=1}^{K}\alpha_{i,k} \cdot \mathbf{h}_{i,k}
\end{gathered}
\label{eq:aggregation_and_attention}
\end{equation}

\noindent where $\mathbf{h}_{i,k}$ is the representation of the $k$-th augmentation neighbor in the augmented representation space, and
$\mathbf{h}_i$ is the new representation of the target sample $\mathbf{x}_i$ obtained via the main encoder $g(\cdot)$, i.e., $\mathbf{h}_i = g(\mathbf{x}_i)$.
$\textbf{W}^{att}\in \mathbb{R}^{n_d \times n_d}$ is the bilinear attention weight matrix, and $\alpha_{i,k}$ corresponds to the attention weight for $\mathbf{h}_{i,k}$.
$\mathbf{h}_{i}^{a}$ and $\mathbf{h}_i$ will then be fused to generate the \textit{cohort-augmented logits} in the final prediction module. 

The incorporation of neighboring vector information can substantially enhance the model's learning capacity, but it may also compromise its disambiguation capability.
As illustrated in Figure \ref{fig: label-aware}, the neighboring samples for $u_i$ may encompass samples from opposite classes.
To mitigate the adverse effects of noisy information from the augmentation neighbors, we build a contrastive objective to separate the negative neighbors in the augmented representation space.
Specifically, we adopt the same transformation layer to obtain representations $\textbf{H}_{i}^{n}$ for the negative neighbors.
Subsequently, we perform a pairwise summation of the target user representation and the negative neighbors' representations. 
These negative-neighbor-augmented user representations 
are then fed into the main output layer to generate $K$ \textit{negative-neighbor-augmented logits} for supervised contrastive learning. The construction of the supervised contrastive loss will be elaborated in Section~\ref{sec:multi_objective_training}. 

\subsection{Prediction Phase}
The target sample representation $\mathbf{h}_{i}$ and the corresponding augmentation neighbor representation $\mathbf{h}_{i}^{a}$ will be fed into an output layer to generate cohort-augmented logit:

\begin{equation}
    \widehat{\mathbf{y}}_i^{a} = \phi_{main}^{a}(\mathbf{h}_i + \mathbf{h}_{i}^{a}), \quad \phi_{main}^{a}:\mathbb{R}^{n_d} \mapsto \mathbb{R}
\label{eq:unlabeled_aug}
\end{equation}

\noindent 
We fuse the two representations via a summation, which empirically is more efficient and stable than concatenation.
The output layer $\phi_{main}^{a}$ will capture the task-specific information from the target sample and augmentation neighbors.
Meanwhile, we can obtain the negative-neighbor-augmented logits $\widehat{\textbf{Y}}_i^{n} \in \mathbb{R}^{n_p \times K}$ by feeding the sum of target user representation and the $K$ negative neighbors' representations into the main output layer.

\subsection{Multi-Objective Joint Training}
\label{sec:multi_objective_training}
We use a joint training scheme for {\mysystem} that combines different objectives in one framework for effective fraud detection.
This helps {\mysystem} to capture intricate fraud behaviors and improve robustness against camouflaged frauds. 

\vspace{1mm}
\noindent
\highlight{Task \#1: Main Supervised Prediction.}
For a fraud detection task, we use BCE as a main objective function:
\begin{equation}
    \mathcal{L}^{a}_{main} = -\frac{1}{N} \sum_{\mathcal{T}_{train}} \mathbf{y}_i log(\sigma (\widehat{\mathbf{y}}_i^{a}))+(1 - \mathbf{y}_i)log(1-\sigma (\widehat{\mathbf{y}}_i^{a}))
\label{eq:main}
\end{equation}
where $\mathbf{y}_i$ and $\widehat{\mathbf{y}}_i^{a}$ are the ground truth label and cohort-augmented prediction logit respectively. $N$ is the number of training samples, and $\sigma(\cdot)$ is the sigmoid function.
With this objective function, {\mysystem} can be trained effectively via gradient-based optimizers.

\vspace{1mm}
\noindent
\highlight{Task \#2: Supervised Contrastive Learning.}
We introduce a \textit{label-aware neighbor separation} mechanism to counter the negative effect of noisy information from augmentation neighbors.
For this objective function, any negative-neighbor-augmented logit $\widehat{\mathbf{y}}_{i,k}^{n}$ in $\widehat{\textbf{Y}}_i^{n}$ should incur greater losses than $\widehat{\mathbf{y}}_i^{a}$, considering the ground truth $\mathbf{y}_i$ of the target sample. Drawing inspiration from prior works~\cite{Yang2022}, we construct the contrastive loss on the prediction logits rather than the cohort-augmented representations, as presented below:

\vspace{-2mm}
\begin{equation}
    \mathcal{L}_{sccl} = -\frac{1}{N}\sum_{\mathcal{T}_{train}} \log \frac{ \exp(-\frac{\dis(\mathbf{y}_i, \widehat{\mathbf{y}}_i^{a})} {\tau})}{\exp(-\frac{\dis(\mathbf{y}_i, \widehat{\mathbf{y}}_i^{a})}{\tau}) + \sum_{k=1}^{K}\exp(-\frac{\dis(\mathbf{y}_i,\widehat{\mathbf{y}}_{i,k}^{n})} {\tau})}
\end{equation}

\noindent where $\dis(\mathbf{y}_i, \widehat{\mathbf{y}}_i^{a})$ represents the distance of the ground-truth logits and cohort-augmented logits, and $\tau$ is the temperature hyperparameter.
With low temperatures, the loss is dominated by closer cohort-augmented logits.
This method enables supervised contrastive learning and prevents the cohort augmented representation from diverging too much from the original burn-in vector. In this manner, the personalized cohort from the burn-in phase remains useful in the augmentation phase.

\vspace{1mm}
\noindent
\highlight{Task \#3: Latent Space Alignment.}
Since cohort identification depends on vector distances between the target sample and candidates, it may deviate slightly from distances in the augmented representation space, potentially reducing the relevance of neighbor information to the current prediction task. To rectify this inconsistency, we introduce auxiliary vector regularization for consistent distances in both vector burn-in and cohort-augmented learning phases:

\begin{equation}
    \mathcal{L}_{align}= \frac{1}{N} \sum_{\mathcal{T}_{train}} \left\| \mathbf{e}_i - \mathbf{h}_i\right\|^2
\end{equation}

The final objective function of {\mysystem} is the summation of the above three tasks and the canonical weight decay regularization:

\begin{equation}
    \mathcal{L}= \mathcal{L}^{a}_{main} + \alpha \mathcal{L}_{sccl} + \beta \mathcal{L}_{align} + \lambda \left \| \theta \right \|^{2}
\end{equation}

\noindent where $\alpha$ and $\beta$ are the hyperparameters that control the weights of the contrastive loss and auxiliary reconstruction loss respectively,
$\theta$ is the set of all model parameters,
and $\lambda$ is the weight regularization strength.

\vspace{1mm}
\noindent
\highlight{Prediction Phase.}
In the final prediction phase, {\mysystem} first obtains the burn-in vector of the target sample $\mathbf{e}_i$ as a query to retrieve the augmentation neighbors.
Next, the augmentation neighbor vectors $\textbf{E}_i^{a}$ are transformed into the augmented representation space $\textbf{H}_{i}^{a}$ and fused with the new representations $\mathbf{h}_i$ of the target sample.
Finally, the augmented representation is fed into the output layer to predict the fraud probability.
\section{Experiments}
\label{sec: experiment}
In our study, we integrate four established fraud detection models with {\mysystem} to assess the effect of the proposed label-aware cohort augmentation on detecting hidden fraudsters.
Particularly, our experiments are devised to answer the key research questions (RQs) as follows:
\begin{itemize} [leftmargin=*] 
\item 
RQ1: Can {\mysystem} improve the existing sequence-based fraud detection model's performance without harming the legitimate users' activities?
\item 
RQ2: How do cohort augmentation and label-aware neighbor separation mechanisms contribute to fraud detection tasks?
\item
RQ3: How does {\mysystem} perform as compared with existing neighbor augmentation approaches? 
\item 
RQ4: How does the size of augmentation and negative neighbor sets impact {\mysystem}'s performance in fraud detection?
\item 
RQ5: How effective is {\mysystem} at unveiling fraud camouflage in graph-based fraud detection methods?
\end{itemize}

\subsection{Experimental Setup}
\subsubsection{Datasets.}
We mainly use two private and one public datasets to evaluate our proposed method. The private datasets are obtained from Shopee~\footnote{\url{https://en.wikipedia.org/wiki/Shopee}}, the largest online e-commerce platform in Southeast Asia. 
Each user is assigned a risk score by a fraud detection model based on their daily activities or purchasing history.
The public dataset is the Amazon~\footnote{\url{https://en.wikipedia.org/wiki/Amazon_(company)}} dataset~\cite{Beutel2013, Jiang2016, Ban2019}, which is widely used for fake review detection. 
Fake review detection is a specific case of fraudulent behavior detection, where each action is a product review.
We compare the performance of the base models with and without {\mysystem} on the three datasets to demonstrate its effectiveness.

\subsubsection*{\textnormal{\textbf{Shopee Dataset}}} The large-scale industrial datasets collected from Shopee adhere strictly to security and privacy protocols.
These datasets consist of historical activity logs originating from the same geographical regions, albeit varying in size, designated as \textbf{Shopee-large} and \textbf{Shopee-small}, spanning a specified timeframe.
Table \ref{table: fraud_data} provides an overview of the dataset statistics. 
For fraud behavior detection, each user is represented by 300 records, each containing details such as actions taken, timestamps, and associated attributes. These actions encompass a spectrum of user interactions within the Shopee platform, including "visit\textit{-}homepage", "refund\textit{-}request", and "password\textit{-}reset\textit{-}request". 
The fraud rate in the table is our sampled fraud rate for the experiments, not the actual fraud rate on Shopee.

We also collected a user-device interaction graph from Shopee, comprising 100k users and 54k devices. This graph dataset serves to validate the effectiveness of {\mysystem} in uncovering fraud camouflage. 
We employ two graph neural networks for fraud detection on this dataset.
Subsequently, we integrate these GNNs with {\mysystem} to observe performance improvements.
To simulate fraudsters' camouflage strategies, we sub-sample the Shopee graph by removing 3758 edges between certain fraudulent accounts and devices. 
Finally, we compare the performance of the two GNNs and their integration with {\mysystem} on this sub-sampled Shopee graph.

\subsubsection*{\textnormal{\textbf{Amazon Dataset}}}
The Amazon dataset comprises diverse product reviews from categories such as electronics, books, CDs, and movies\footnote{\url{https://jmcauley.ucsd.edu/data/amazon/}}. Extensive use case analyses conducted in previous studies \cite{Beutel2013, Jiang2016, Ban2019} have revealed patterns of fraudulent activities, including repeated purchases of identical items and the rapid submission of reviews. 
Utilizing the filtering algorithms in \cite{Zhang2020, Dou2020}, users are labeled based on their helpful vote percentages, with those below 20\% as fake users and those above 80\% as normal users.

\begin{table}[t]
  \centering
  \small
  \caption{Statistics of datasets used in the fraud detection task. The \textit{Pos. rate} is the ratio of fraud users.}
  \setlength{\tabcolsep}{3pt}
    \begin{tabular}{lccccc}
    \toprule
    \textbf{Datasets} & \#Users & \#Action types & \#Actions & \#Fields & Pos. rate \\
    \midrule
    Shopee-small & 16k & 653 & 4.8M & 11 & 1\% \\
    \midrule
    Shopee-large & 1.05M & 653 & 315M & 11 & 1\% \\
    \midrule
    Amazon & 18k & 719820 & 13.3M & 3 & 6.86\% \\
    \bottomrule
    \end{tabular}%
    \vspace{-2mm}
  \label{table: fraud_data}%
\end{table}%

\subsubsection{Baseline Methods}
We integrate {\mysystem} with four sequence-based fraud detection models and compare its performance to two existing neighbor augmented learning methods.
Additionally, we utilize two graph neural networks to validate the effectiveness of {\mysystem} in uncovering camouflaged fraud.
Ablation tests are also conducted to show the efficacy of the techniques proposed in {\mysystem}.

\vspace{1mm}
\noindent 
\highlight{Sequence-based fraud detection base models.} 
(1) \textbf{BiLSTM}~\cite{Schuster1997} applies bidirectional LSTM to capture the dependency information among sequential data for fraud detection.
(2) \textbf{LIC Tree-LSTM}~\cite{Liu2020} builds a behavior tree for the user's behavioral sequence and studies fraud transaction detection with local intention calibration.
(3) \textbf{HEN}~\cite{Zhu2020} uses field-level extractor and action-level extractor to hierarchically learn users' intentions and identify high-risk behavior sequences.
(4) \textbf{MINT}~\cite{Xiao2023} employs a time-aware behavior graph to model user's behavioral data and utilizes a multi-view graph neural network to detect fraudulent behaviors.

\vspace{1mm}
\noindent
\highlight{Graph-based fraud detection base models.} 
(1) \textbf{GraphSAGE}~\cite{Hamilton2017} performs inductive learning on graphs by sampling and aggregating node representations.
(2) \textbf{GeniePath}~\cite{Liu2019} integrates LSTMs into GCN layers to adaptively aggregate neighborhood information.

\vspace{1mm}
\noindent 
\highlight{Existing neighbor augmentation methods.} (1) \textbf{GRASP}~\cite{Zhang2021} divides samples into different clusters and uses the cluster representations to enrich sample representations. (2) \textbf{CSRM}~\cite{Wang2019d} dynamically retrieves different neighbors' information during model training for augmented learning.

\vspace{1mm}
\noindent
\highlight{Ablations.} \textbf{{\mysystem}$_{BI}$} is a variant of {\mysystem} that adopts a vanilla cohort augmentation architecture with a vector burn-in mechanism. \textbf{{\mysystem}$_{LA}$} is a variant of {\mysystem} that only introduces the label-aware neighbor separation mechanism.

\subsubsection{Implementation Details}
In our implementation, we utilize PyTorch for all models and Milvus~\cite{Wang2021b} to store embedding vectors.
For the datasets, the split ratio of the training/validation/testing set is 60\%/20\%/20\%.
We employ the Adam optimizer and initialize model parameters with Xavier initialization.
Early stopping is applied if the validation performance does not improve for 20 epochs.
The settings of the fraud detection task follow the practice in~\cite{Liu2020, Xiao2023}. 
The embedding dimension of users' behavioral action is set to 64, the batch size to 128, the learning rate to 0.0001, and the maximum behavioral sequence length to 300.
The regularization strength is set to $1e$-$5$.
The number of nearest neighbors $K$ is also set to five for all models and datasets. 

We used grid search to optimize three key hyperparameters for {\mysystem} with different base models: the contrastive loss weight $\alpha$, the auxiliary reconstruction loss weight $\beta$, and the temperature $\tau$.
These hyperparameters control the importance of the contrastive objective, the balance between cohort-augmented learning and latent space alignment, and the focus on close or far-away cohort neighbors, respectively. 
We search $\alpha$ in (0.00001, 0.0001, 0.001, 0.01, 0.1, 1.0), $\beta$ in (0.0000001, 0.000001, 0.00001, 0.0001, 0.001), and $\tau$ in (0.1, 0.3, 1.0, 3.0, 10.0).
These ranges were chosen to cover a wide spectrum of possible values for different fraud detection models. We found that $\tau = 1$ was the best value for all models.
Table~\ref{table: hyper} shows the values of $\alpha$ and $\beta$ selected for {\mysystem} with various base models and datasets.

\begin{table}[t]
  \centering
  \small
  \caption{The hyperparameters selected for {\mysystem} on different fraud detection methods and three distinct datasets.}
  \setlength{\tabcolsep}{1pt} 
    \begin{tabular}{@{}l *{12}{c} @{}}
    \toprule
    \multirow{3}{*}{\textbf{Datasets}} & 
    \multicolumn{2}{c}{\multirow{2}{*}{\textbf{BiLSTM}}} &
    \multicolumn{2}{c}{\textbf{LIC}} &
    \multicolumn{2}{c}{\multirow{2}{*}{\textbf{HEN}}} &
    \multicolumn{2}{c}{\multirow{2}{*}{\textbf{MINT}}} \\
    & & & \multicolumn{2}{c}{\textbf{Tree-LSTM}} & & & & & \\
    & $\alpha$ & $\beta$ & $\alpha$ & $\beta$  & $\alpha$ & $\beta$ & $\alpha$ & $\beta$ & \\
    \midrule
    \multirow{1}{*}{\makecell[l]{Shopee-small}} & 0.01 & 0.00001 & 0.01 & 0.00001 & 0.001 & 0.00001 & 0.001 & 0.00001 \\ 
    \midrule
    \multirow{1}{*}{\makecell[l]{Shopee-large}} & 0.001 & 0.000001 & 0.001 & 0.000001 & 0.001 & 0.000001 & 0.001 & 0.00001 \\
    \midrule
    \multirow{1}{*}{\makecell[l]{Amazon}} & 0.01 & 0.00001 & 0.01 & 0.00001 & 0.01 & 0.00001 & 0.001 & 0.00001 \\ 
    \bottomrule
    \end{tabular}%
    \vspace{-2mm}
  \label{table: hyper}%
\end{table}%

\begin{table*}[t]
  \centering
  \caption{Experimental results of different sequence-based fraud detection methods with (w/) and without (w/o) {\mysystem} on the three datasets. The best results are highlighted in boldface.}
    \begin{tabular}{llccc | ccc | ccc | ccc}
    \toprule
    \multirow{2}{*}{\textbf{Datasets}} & 
    \multirow{2}{*}{\textbf{Metric}} & 
    \multicolumn{3}{c}{\textbf{BiLSTM}} & 
    \multicolumn{3}{c}{\textbf{LIC Tree-LSTM}}  & 
    \multicolumn{3}{c}{\textbf{HEN }} & 
    \multicolumn{3}{c}{\textbf{MINT}} \\
    &  & w/o & w/ & Imprv. & w/o & w/ & Imprv. & w/o  & w/ & Imprv. & w/o & w/ & Imprv. \\
    \midrule
    \multirow{2}{*}{\makecell[l]{Shopee-small}} &
    AUC & 0.8567 & \textbf{0.8678} & 1.30\% & 0.8489 & \textbf{0.8577} & 1.04\% & 0.8611 & \textbf{0.8730} & 1.38\% & 0.8892 & \textbf{0.8971} & 0.89\% \\ &
    R@P$_{0.9}$ & 0.4789 & \textbf{0.5103} & 6.56\% & 0.4201 & \textbf{0.4578} & 8.97\% & 0.4737 & \textbf{0.4983} & 5.19\% & 0.5102 & \textbf{0.5387} & 5.59\% \\
    \midrule
    \multirow{2}{*}{\makecell[l]{Shopee-large}} &
    AUC & 0.8778 & \textbf{0.8869} & 1.04\% & 0.8723 & \textbf{0.8822} & 1.13\% & 0.8805 & \textbf{0.8898} & 1.06\% & 0.9087 & \textbf{0.9175} & 0.97\% \\ &
    R@P$_{0.9}$ & 0.5267 & \textbf{0.5536} & 5.11\% & 0.4149 & \textbf{0.4378} & 5.52\% & 0.5304 & \textbf{0.5673} & 6.96\% & 0.5735 & \textbf{0.5982} & 4.31\% \\
    \midrule
    \multirow{2}{*}{\makecell[l]{Amazon}} &
    AUC & 0.8788 & \textbf{0.8917} & 1.47\% & 0.8656 & \textbf{0.8871} & 2.48\% & 0.8811 & \textbf{0.8967} & 1.77\% & 0.9134 & \textbf{0.9245} & 1.22\% \\ &
    R@P$_{0.9}$ & 0.2231 & \textbf{0.2513} & 12.64\% & 0.1938 & \textbf{0.2374} & 22.50\% & 0.2411 & \textbf{0.2578} & 6.93\% & 0.3135 & \textbf{0.3317} & 5.81\% \\
    \bottomrule
    \end{tabular}%
  \label{table: fraud results}%
\end{table*}%

\subsubsection{Evaluation Metrics}
For fraud detection tasks, we evaluate all methods with two widely used metrics: AUC (Area Under ROC) and R@P$_k$. AUC signifies the probability that positive cases' predictions are ranked ahead of negative cases. R@P$_k$ represents the recall rate when precision equals $k$. 
In our fraud detection scenario, the goal is to identify as many suspicious entities as possible while not disrupting the regular operations of benign entities. We set $k$ to 0.9 for all experiments. 
A superior cohort-augmented learning framework is expected to achieve higher AUC values than base models and notably enhance the recall rate over base models when precision values are set to 0.9.

\subsection{Effectiveness of Cohort Augmentation(RQ1)}
Table \ref{table: fraud results} shows the fraud detection performance of four state-of-the-art baselines with and without {\mysystem}.
{\mysystem} improves both the AUC and R@P$_{0.9}$ of the base models significantly.
On the Shopee datasets, the AUC increases by at least 0.89\% and R@P$_{0.9}$ by more than 4.31\%. On the Amazon dataset, the AUC rises by at least 1.22\% and R@P$_{0.9}$ by 5.81\%. 
Notably, the substantial enhancement in R@P$_{0.9}$ indicates that {\mysystem} aids in identifying more fraudsters who escape the detection of base models.
These fraudsters mimic the behavior of normal users to avoid being caught.
{\mysystem} uses the task-specific embedding vectors of the base models to create personalized implicit user cohorts, linking camouflaged fraudsters implicitly to other users who are controlled by the same individuals or groups.
Unlike conventional GNN-based approaches vulnerable to intentionally masked identity features of fraudsters, our cohort identification methods rely on learned user behavioral representations to detect the camouflaged fraudsters effectively.
For example, in Shopee, users with similar behavioral sequences may have similar user representations learned by base models. 
However, the base models cannot adequately separate fraud from normal users due to the presence of many camouflaged fraudsters.
{\mysystem} connects camouflaged fraudsters to labeled fraud users and separates them from similar normal users using a label-aware neighbor separation mechanism, which enables better discrimination between suspicious and normal behaviors.

\begin{figure}
\centering
\includegraphics[scale=0.25]{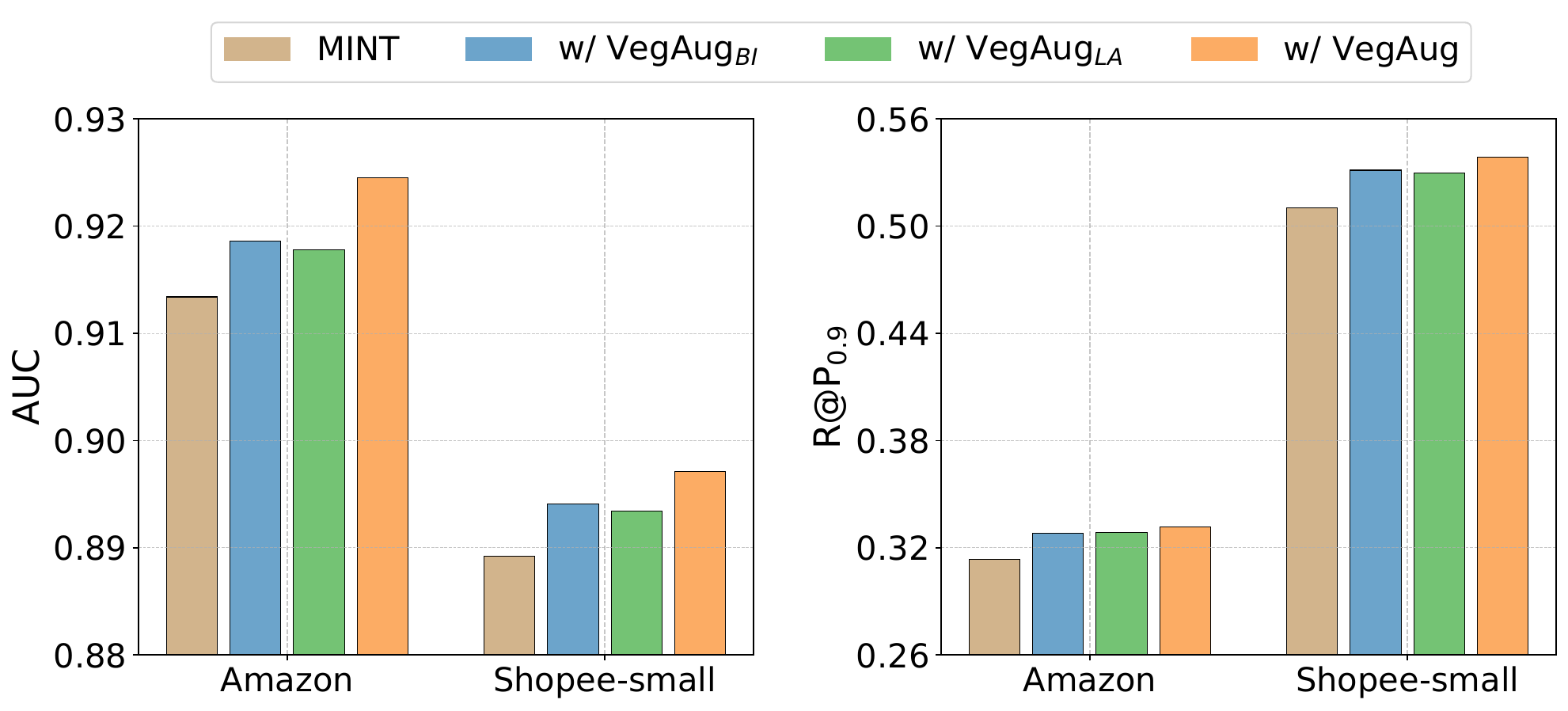}
\caption{Performance comparison of {\mysystem} and its two variants in fraud detection tasks.}
\label{fig: ablation}
\end{figure}

\subsection{Ablation Studies (RQ2)}
\label{subsec: ablation studies}
We conduct ablation studies on {\mysystem} to demonstrate the impact of vector burn-in cohort augmentation and label-aware neighbor separation on its performance.

\subsubsection{Vector Burn-in for Cohort Augmentation}
Figure \ref{fig: ablation} demonstrates that the vector burn-in technique alone substantially improves base model performance.
This indicates that the learning-based cohort approach is effective for fraud detection. 
By constructing event-based networks for each user, we can leverage the similarities among users to infer their labels. 
Moreover, we can link unseen suspicious activities to existing samples, and obtain more relevant information for prediction. 
Fraud detection is a binary task, but fraud behavioral patterns are diverse and complex. 
The user behavioral representations learned from any base models will not form two clear clusters. 
Instead, they will form many clusters of varying sizes, especially for user accounts operated by the same person or group. 
Normal users’ representations are usually more distinct from each other. Therefore, if we construct a personalized cohort for each user, it will mainly affect fraud users and make their clusters more compact. 
Fraud users not affiliated with any cluster are connected to other fraud users who are similar to them. 
This has minimal impact on normal users, who are less likely to be operated by a common entity, and whose cohort-augmented representation will not change much. As a result, we can detect more fraudulent users without compromising the performance of normal users.

\subsubsection{Label-aware Neighbor Separation Mechanism}  
Figure~\ref{fig: ablation} demonstrates a significant enhancement in both AUC and R@P$_{0.9}$ metrics across both datasets upon integrating MINT with {\mysystem}$_{LA}$. 
While cohort augmentation fosters connections between various fraudsters and enhances learned representations, it may inadvertently link adept fraudsters with normal users exhibiting similar behavioral patterns. 
This scenario introduces considerable noise through direct message passing between fraudulent and normal users, thereby notably compromising model performance. 
To mitigate this issue, the label-aware neighbor separation mechanism deliberately distances these challenging negative neighbors during cohort augmentation via a supervised contrastive loss. 
Given the original base models' inability to discern target users from their difficult negative counterparts, the contrastive objective function serves to recalibrate the learning bias, prioritizing the treatment of these negative pairs and effectively separating them within the augmented learning space. 
This calibration is achieved through the construction of the contrastive loss on the augmented prediction logits, primarily influencing the output layers while also exerting a lesser influence on the encoder training.

\subsection{Comparison with Existing Neighbor Augmentation Methods (RQ3)}
\begin{figure}
\centering
\includegraphics[scale=0.25]{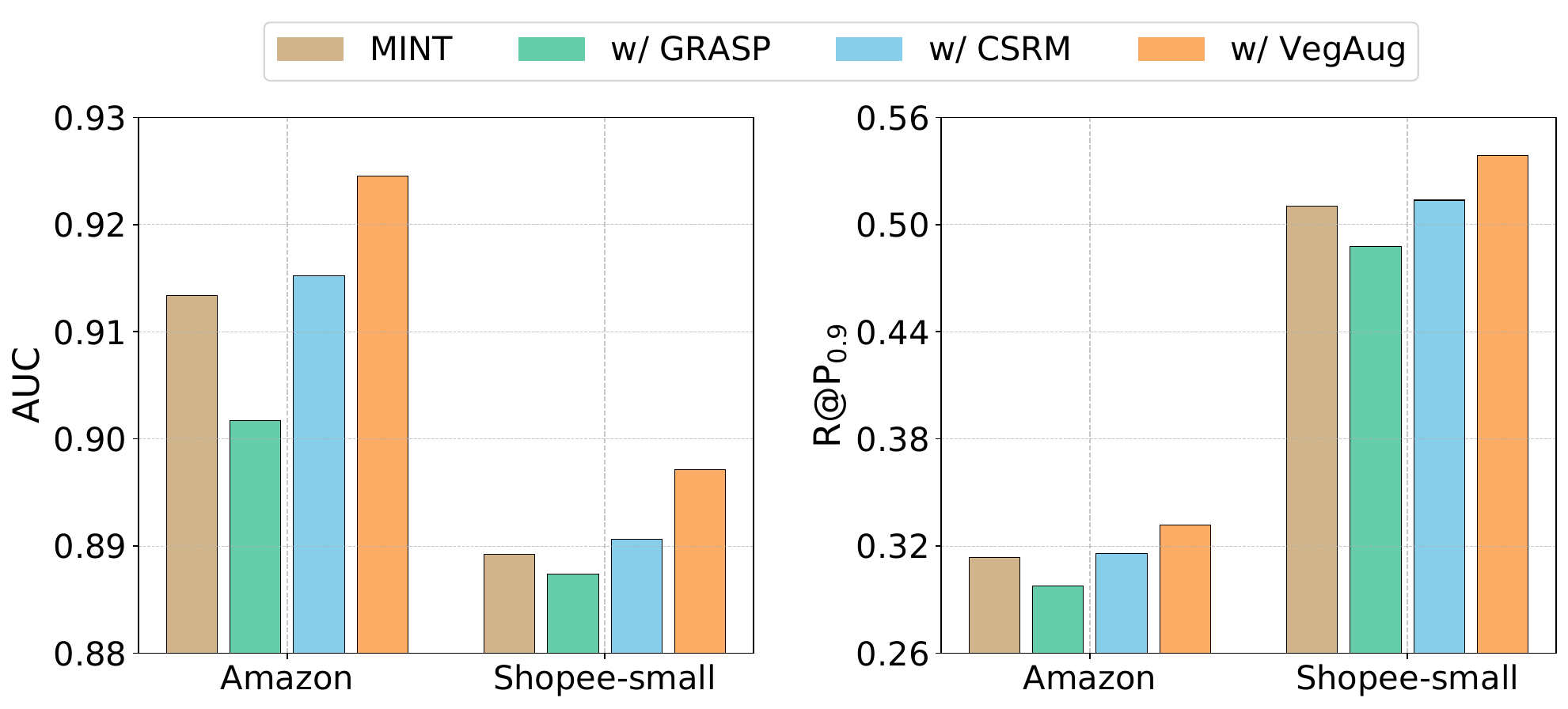}
\caption{Comparison of the effectiveness of {\mysystem}, GRASP, and CSRM in enhancing the discriminative ability of the existing fraud detection models.
}
\vspace{-2mm}
\label{fig: old_aug}
\end{figure}

In this subsection, we compare the effectiveness of {\mysystem} with two existing methods that use neighborhood information to enhance target users: GRASP~\cite{Zhang2021} and CSRM~\cite{Wang2019d}. 
Figure \ref{fig: old_aug} illustrates the performance comparisons. 
GRASP clusters the samples into groups based on the task-specific vectors using K-Means, and uses the centroid vectors as group representations. 
It constructs a KNN graph for different groups, with each user aggregating information from its group and other connected groups. 
However, GRASP functions as a three-layer graph neural network and passes excessive information from the neighbors to the target users, leading to severe over-smoothing. 
Moreover, GRASP compromises users’ personalized behavioral patterns and performs worse than the base model on both datasets. 
CSRM dynamically retrieves the nearest neighbors for each user, demonstrating more effective neighbor augmentation than GRASP. 
However, CSRM encounters challenges in accurately utilizing the neighbors’ information in the early stages of training, which can lead to erroneous results. 
It also fails to identify the most useful augmentation neighbors for the target user. Our {\mysystem} solves these problems by introducing a new vector burn-in technique for more precise cohort identification and a label-aware contrastive neighbor separation mechanism for mitigating the negative effect of noisy information. 
From Figure \ref{fig: old_aug}, we can see that {\mysystem} outperforms and is more consistent than both the previous neighbors-augmented learning methods in fraud detection.

\begin{figure}
\centering
\includegraphics[scale=0.25]{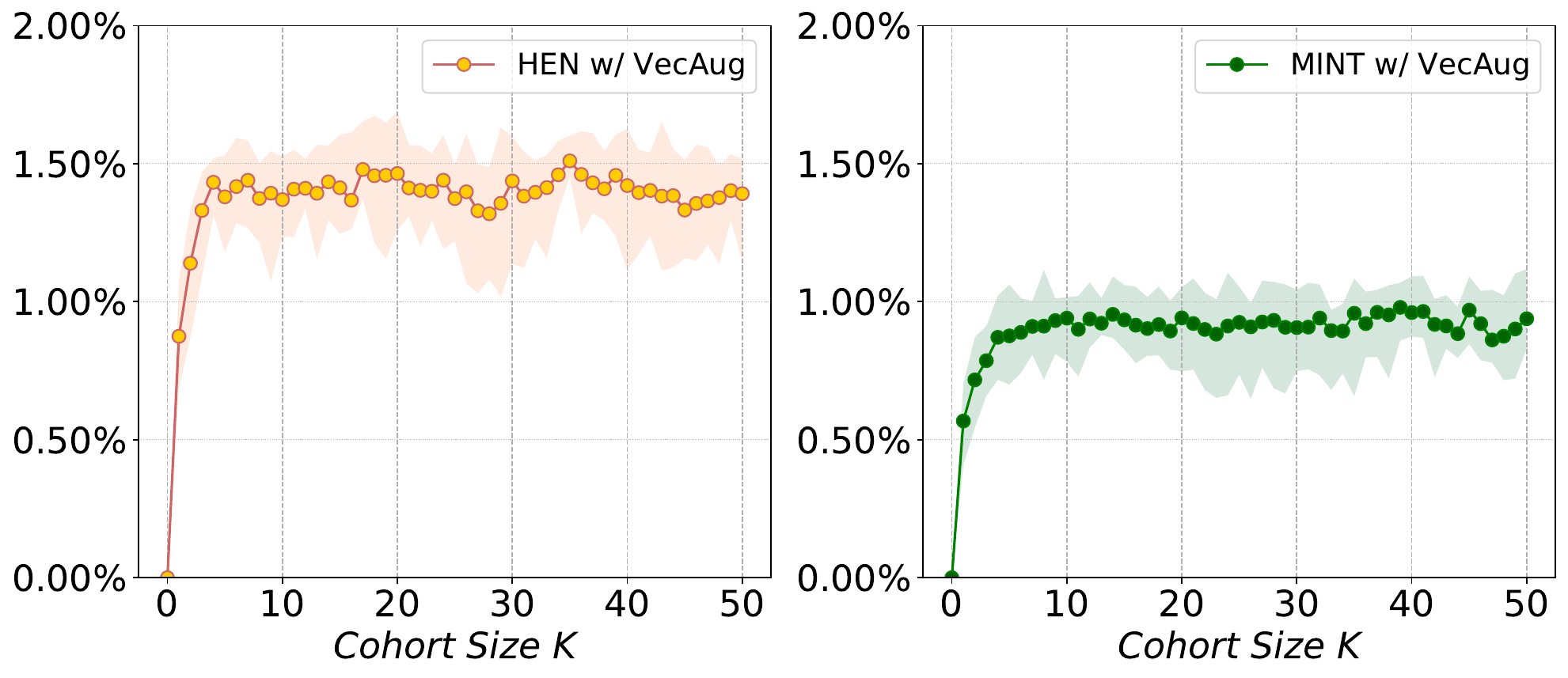}
\caption{Performance improvements of HEN and MINT on Shopee-small after integration with {\mysystem}, depicted in AUC across various cohort sizes.}
\vspace{-2mm}
\label{fig: kNNsize}
\end{figure}

\subsection{Influence of Cohort Size (RQ4)}
In this section, we investigate the impact of cohort size variation on the effectiveness of {\mysystem}. The results obtained from two fraud detection models on the Shopee-small dataset are illustrated in Figure \ref{fig: kNNsize}. 
It is evident that the performance of the proposed label-aware cohort-augmented learning framework remains highly robust across varying cohort sizes once the cohort size exceeds a certain threshold. 
Specifically, AUC values demonstrate an increasing trend with cohort size, plateauing when $K$ reaches four. With a smaller cohort, meaningful information is aggregated from training samples to enrich the representation of the target sample. 
However, as the cohort size increases, both valuable and noisy information is introduced to the target users, resulting in performance gains that saturate at $K$=4. 
Initially, most information learned from augmentation neighbors closely aligns with the target users, thereby enhancing their representation. 
Yet, with an increased cohort size, the likelihood of encountering augmentation neighbors with opposite labels to the target users rises, potentially compromising the effectiveness of newly augmented neighbors. 
However, the implementation of a label-aware neighbor separation scheme helps mitigate the adverse effects of noisy information introduced by the negative neighbors. 
When the impacts of negative neighbors and the label-aware neighbor separation scheme are comparable, the performance of {\mysystem} remains relatively stable across varying cohort sizes. In summary, our proposed {\mysystem} effectively enhances the learning capacity of base models and exhibits robust performance, irrespective of cohort size variations.

\subsection{Studies of Unveiling Camouflaged Fraud (RQ5)}
To validate the {\mysystem}'s effectiveness in unveiling the camouflage of frauds, we simulate fraudsters' camouflage strategies by selectively removing 3758 edges between certain fraudulent accounts and devices in the Shopee-graph dataset.
We then compare the performance of GraphSAGE and GeniePath with and without {\mysystem} integration on both the original Shopee-graph and the sub-sampled Shopee-graph datasets.
The comparison results in Table~\ref{table: camouflage} reveal that both GraphSAGE and GeniePath experience severe degradation when fraudster-device connections are deliberately removed.
However, the integration of {\mysystem} significantly enhances the effectiveness of GraphSAGE and GeniePath on the sub-sampled Shopee-graph. 
Specifically, {\mysystem} enables GraphSAGE to recover 86.3\% (3242/3758) and GeniePath to recover 82.9\% (3115/3758) of the deliberately removed connections.
This leads to R@P$_{0.9}$ improvements of 47.61\% and 40.02\%, highlighting {\mysystem}’s strength in identifying more fraudulent users and thus unveiling the camouflaged fraud.
Furthermore, the comparison between the base GNN models and their integration with {\mysystem} demonstrates that {\mysystem} can also augment the learning capabilities of GNN-based models in fraud detection.

\begin{table}[t]
  \centering
  \caption{GraphSAGE and GeniePath performance with and without {\mysystem} integration on Shopee-graph and sub-sampled Shopee-graph.}
    \begin{tabular}{lccc | cc }
    \toprule
    \multirow{2}{*}{\textbf{Models}} & 
    \multicolumn{1}{c}{\makecell[c]{\textbf{Datasets}}} & 
    \multicolumn{2}{c}{\makecell[c]{Shopee-graph}}  & 
    \multicolumn{2}{c}{\makecell[c]{Sub-sampled \\ Shopee-graph}} \\
    & \textbf{Metric} & AUC & R@P$_{0.9}$ & AUC & R@P$_{0.9}$ \\
    \midrule
    \multirow{3}{*}{\textbf{GraphSAGE}} &
    w/o & 0.8536 & 0.3417 & 0.8051 & 0.2243 \\ &
    w/  & \textbf{0.8623} & \textbf{0.3638} & \textbf{0.8486} & \textbf{0.3311} \\ &
    Imprv. & 1.02\% & 6.47\% & 5.40\% & 47.61\% \\ 
    \midrule
    \multirow{3}{*}{\textbf{GeniePath}} &
    w/o & 0.8714 & 0.3756 & 0.8194 & 0.2541 \\ &
    w/ & \textbf{0.8789} & \textbf{0.3887} & \textbf{0.8573} & \textbf{0.3558} \\ &
    Imprv. & 0.86\% & 3.49\% & 4.63\% & 40.02\% \\ 
    \bottomrule
    \end{tabular}%
  \label{table: camouflage}
\end{table}%

\subsection{Efficiency of {\mysystem} with Integrating Vector Database}
\begin{figure}
\centering
\includegraphics[scale=0.29]{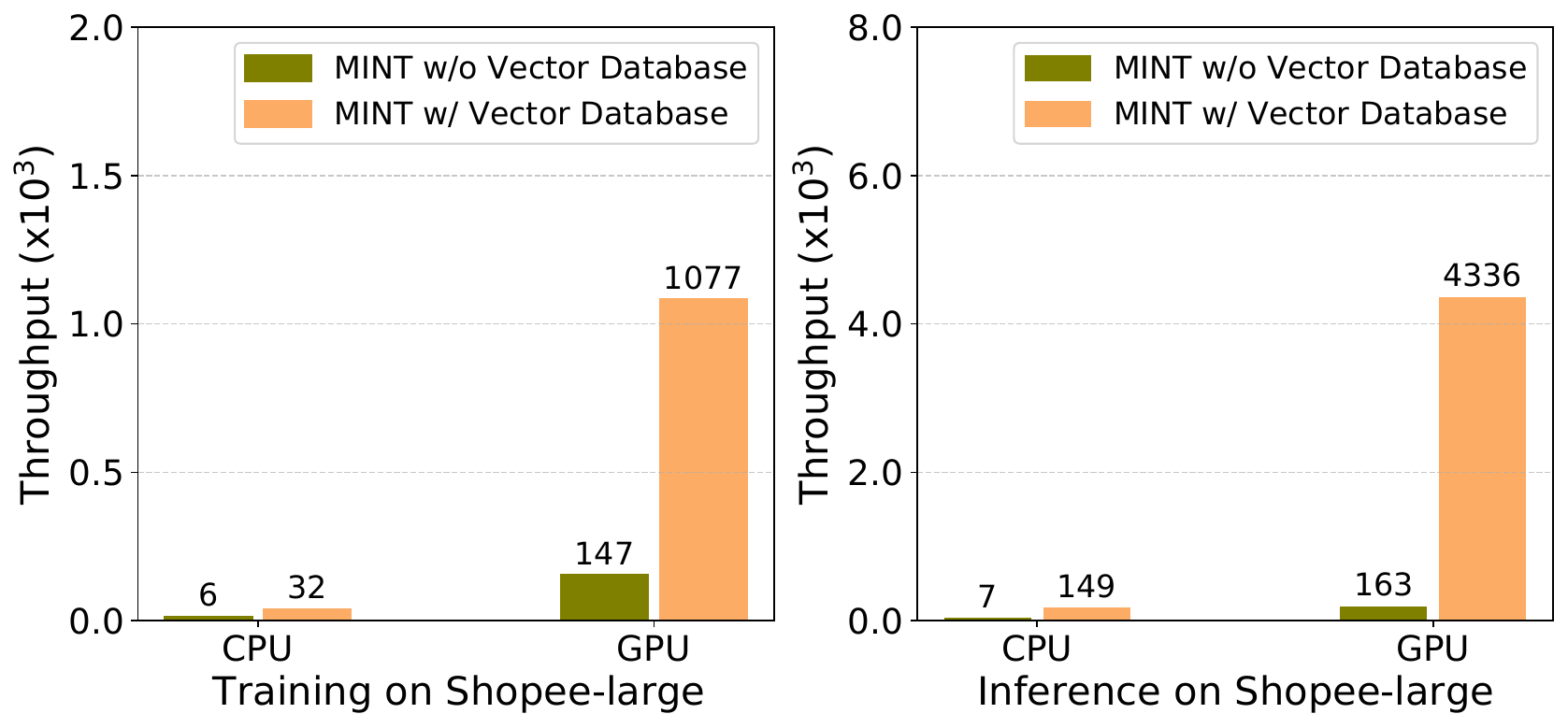}
\vspace{-1mm}
\caption{Throughput comparison of {\mysystem} with and without vector database (with index HNSW).}
\vspace{-2mm}
\label{fig: throughput}
\end{figure}

We assess the training and inference efficiency of {\mysystem} with and without Milvus in this subsection, integrating with MINT on the Shopee-large dataset. 
We performed experiments over 20 epochs for both cases and measured the throughput (number of training/inference tuples processed per second) for both CPU and GPU. 
To ensure fair comparisons, we maintained a consistent batch size of 512 across all experiments to maximize the GPU computational capacity. 
For Milvus, we employed HNSW as the index for vector search, known for its high accuracy and efficient indexing capabilities. 
In contrast, for the scenario without a vector database, we utilized PyTorch's built-in functions for cohort neighbor searches.
The throughput comparison results, as depicted in Figure~\ref{fig: throughput}, demonstrate that {\mysystem} exhibits significantly improved efficiencies when integrated with a vector database compared to {\mysystem} operating without one.
Specifically, the solution of {\mysystem} with Milvus achieves significantly higher training and inference efficiency, with speedups of up to 21.3x and 26.6x of CPU and GPU respectively, and incurs no extra infra investment.
Our experimental results highlight {\mysystem}'s seamless integration with existing fraud detection methods and its effective combination with vector databases, providing a more efficient approach to fraud detection.

\section{Related Work}
\label{sec: related}
Fraud detection tasks analyze users’ behavioral data and transaction history to identify fraudulent activities, such as fake reviews, brushing, fraud payments, etc. 
Existing fraud detection systems fall into two categories: graph-based and sequence-based. 
Graph-based methods~\cite{Liu2018, Wang2019b, Wang2021c, Cheng2020a, Dou2020, Dou2020b, Liu2021b, Zhang2023b, Huang2022, Shi2022, Yu2023} use predefined features like devices, IPs, and purchase patterns to connect users, but these features may not reflect behavioral similarity and are easy to manipulate by fraudsters ~\cite{Dou2020, Wang2021c}.
In contrast, sequence-based methods use recurrent neural networks (RNNs)~\cite{Wang2017, Li2018, Branco2020, Liu2020, Lin2021, Wang2023} or attention-based mechanisms~\cite{Guo2019b, Zheng2019, Zhu2020, Liu2020, Liu2021} to capture temporal and sequential behavior patterns, which are more robust against manipulation.
For example, some studies utilize behavior trees or Tree-LSTM~\cite{Liu2020, Liu2021} to extract user intentions from sequences, while others propose hierarchical explainable networks~\cite{Zhu2020} to predict payment risks using critical historical behavioral sequences.
More recent works~\cite{Wang2023} design a general user behavior modeling framework to pre-train users’ behavior data and learn the robust representation for fraud detection.
However, sequence-based methods often neglect user interactions, which are crucial for detecting over-consistent behavior patterns in real-world fraud scenarios.
To address this, we propose a learning-based cohort augmentation method, which constructs personalized cohorts for each user based on the relevance of their behavioral sequences. 
In this manner, we can capture the subtle and dynamic patterns of fraudulent behaviors, and enhance the representation of each user.

\vspace{-2mm}
\section{Conclusions}
\label{sec: conclusion}
This paper introduces {\mysystem}, a label-aware cohort-augmented learning framework designed to enhance the fraud detection model’s capability in revealing camouflaged frauds.
Unlike conventional neighbor augmentation approaches,
{\mysystem} seamlessly integrates with existing models, offering a versatile solution to enrich user representations.
{\mysystem} incorporates a task-specific vector burn-in technique, enabling it to select augmentation neighbors and negative neighbors automatically for each training user. 
To refine the aggregated cohort information, {\mysystem} employs a label-aware neighbor separation mechanism, effectively distancing the negative neighbor representations in the augmented space. 
It addresses the limitations of conventional fraud detection methods by using learned behavior similarity instead of identity features to build connections between users, effectively handling noise and outliers, and explicitly considering label information during augmentation.
Comprehensive experiments and evaluations have demonstrated {\mysystem}'s effectiveness and advantages in real-world fraud detection tasks. 
In summary, {\mysystem} provides a robust approach to enhance the distinguishing ability of existing fraud detection models, providing a promising strategy for detecting camouflaged fraudsters.

\begin{acks}
This research is supported by Shopee Singapore Private Limited, the Singapore Economic Development Board, and the Singapore Ministry of Education Academic Research Fund Tier 3 under MOE’s official grant number MOE2017-T3-1-007.
The work of Meihui Zhang is supported by the National Natural Science Foundation of China (62072033).
\end{acks}



\bibliographystyle{ACM-Reference-Format}
\balance
\bibliography{reference}

\clearpage
\appendix
\nobalance

\clearpage
\appendix

\section*{Appendices}
\section{PSEUDO-CODE OF {\mysystem}}
\label{sec:PSEUDO_CODE}
Algorithm \ref{alg: burnin} provides a detailed pseudocode description of the vector burn-in technique employed by {\mysystem}. This technique represents a fundamental aspect of {\mysystem}'s approach, effectively separating the optimization of embedding vectors from their utilization in subsequent cohort augmentation. Additionally, {\mysystem} stores the label information of training samples, a crucial element that plays a significant role in the subsequent negative neighbor separation process, enhancing its overall effectiveness.

On the other hand, Algorithm \ref{alg: augmentation} outlines the procedure for cohort identification and subsequent cohort-augmented training within the {\mysystem} framework. 
To enhance clarity, certain steps, including contrastive loss calculation, and vector regularization, have been omitted for simplicity.
During this process, it's essential to note that the parameters of auxiliary encoder $g_{aux}(\cdot)$ and the output layer used in the vector burn-in phase will remain fixed. This deliberate decision ensures the preservation of task-specific relevance between the target sample and the neighboring vectors, thereby facilitating the augmentation and separation of the representations between neighbors and target samples.
Furthermore, by retaining well-trained neighborhood information, the augmented-training process becomes significantly more manageable in comparison to dynamic neighbor identification and augmentation methods. This design choice not only promotes computational efficiency but also contributes to the stability and effectiveness of {\mysystem} in enhancing fraud detection.

\begin{algorithm}[t]
  \caption{Vector Burn-in in \textit{\mysystem}}
  \label{alg: burnin}
  \DontPrintSemicolon
  \KwIn{Training data $\mathcal{T}_{\text{train}}$; Auxiliary encoder $g_{aux}(\cdot)$;}
  \KwOut{Well-trained auxiliary encoder $g_{aux}(\cdot)$; Vector pool $\mathcal{E}_{pool}$.}
  \For{each epoch}{
    \For{each sample $u_i \in \mathcal{T}_{\text{train}}$}{
      Encode $\mathbf{x}_i$ to generate vectors: $\mathbf{e}_i = g_{aux}(\mathbf{x}_i)$\;
      Generate prediction value: $\widehat{\mathbf{y}}_i = \phi_{aux}(\mathbf{e}_i)$\;
      Calculate the loss: $\mathcal{L}_{aux} = \text{loss}(\widehat{\mathbf{y}}_i, \mathbf{y}_i)$\;
      Calculate the gradients: $\nabla\mathcal{L}_{aux} = \frac{\partial\mathcal{L}_{aux}}{\partial\mathbf{w}_{aux}}$\;
      Update the model parameters: $\mathbf{w}_{aux} = \mathbf{w}_{aux} - \alpha \nabla\mathcal{L}_{aux}$\;
    }
  }
  \For{each sample $u_i \in \mathcal{T}_{\text{train}}$}{
    Perform a forward pass: $\mathbf{e}_i = g_{aux}(\mathbf{x}_i)$\;
    Store the vectors and their corresponding labels in the vector database $\mathcal{E}_{pool}$: $\mathcal{E}_{pool}.add(\mathbf{e}_i, \mathbf{y}_i)$\;
  }
  \Return{$g_{aux}(\cdot)$, $\mathcal{E}_{pool}$}\;
\end{algorithm}
\vspace{-2mm}

\begin{algorithm}[t]
  \caption{Cohort-augmented Training in \textit{\mysystem}}
  \label{alg: augmentation}
  \DontPrintSemicolon
  \KwIn{Training data $\mathcal{T}_{\text{train}}$; Vector pool $\mathcal{E}_{pool}$; Well-trained auxiliary encoder $g_{aux}(\cdot)$; Number of neighbors $K$;}
  \KwOut{Well-trained cohort augmentation model;}
  \For{each epoch}{
    \For{each sample $u_i \in \mathcal{T}_{\text{train}}$}{
    Perform a forward pass: $\mathbf{e}_i = g_{aux}(\mathbf{x}_i)$\;
    Search top-$K$ similar cohort neighbors from $\mathcal{E}_{pool}$ for $u_i$, obtain $\mathbf{E}_i^{a}$: $\mathbf{E}_i^{a} = \text{search}(\mathbf{e}_i, \mathcal{E}_{pool})$\;
    Search top-$K$ similar vectors from $\mathcal{E}_{pool}$ for $u_i$ with different labels, obtain $\mathbf{E}_i^{n}$: $\mathbf{E}_i^{n} = \text{search}(\mathbf{e}_i, \mathcal{E}_{pool}, \mathbf{y_e} \neq \mathbf{y_u})$\;
    Encode $\mathbf{x}_i$ to generate representations: $\mathbf{h}_i = g(\mathbf{x}_i)$\;
    Generate cohort-augmented logits: $\widehat{\mathbf{y}}_i^{a} = f(\mathbf{h}_i; \mathbf{e}_i; \textbf{E}_i^{a}; \mathbf{E}_i^{n}; \mathbf{W}_f)$ \;
    Calculate the loss: $\mathcal{L}^{a}_{main} = \text{loss}(\widehat{\mathbf{y}}_i^{a}, \mathbf{y}_i)$\;
    Compute the gradients: $\nabla\mathcal{L}^{a}_{main} = \frac{\partial\mathcal{L}^{a}_{main}}{\partial\mathbf{w}^{a}_{main}}$\;
    Update the model parameters: $\mathbf{w}^{a}_{main} = \mathbf{w}^{a}_{main} - \alpha \nabla\mathcal{L}^{a}_{main}$\;
    }
  }
  \Return{Well-trained cohort augmentation model}\;
\end{algorithm}

\section{More Experimental Results}
\label{sec: more_results}
We also conducted experiments on two more methods widely compared recently, i.e., IHGAT~\cite{Liu2021} and FATA-Trans~\cite{Zhang2023b} to verify the effectiveness of {\mysystem}.
The results in Table~\ref{table: more_results} show that {\mysystem} considerably improves the detection performance of these two representative methods across all three datasets, with AUC (1\% increase is considered significant~\cite{Liu2021, Zhang2023b}) 
and R@P$_{0.9}$ improvements up to 1.53\% and 6.83\%, respectively.
These new results further validate the effectiveness of our cohort augmentation for fraud detection.

\begin{table}[t]
  \centering
  \caption{Experimental results of IHGAT and FATA-Trans with and without {\mysystem}.}
  \setlength{\tabcolsep}{1.5pt}
    \begin{tabular}{llccc | ccc }
    \toprule
    \multirow{2}{*}{\textbf{Datasets}} & 
    \multirow{2}{*}{\textbf{Metric}} & 
    \multicolumn{3}{c}{\textbf{IHGAT}}  & 
    \multicolumn{3}{c}{\textbf{FATA-Trans}} \\
    &  & w/o & w/ & Imprv. & w/o & w/ & Imprv. \\
    \midrule
    \multirow{2}{*}{\makecell[l]{Shopee-small}} &
    AUC & 0.8571 & 0.8725 & 1.18\% & 0.8772 & 0.8875 & 1.17\% \\ &
    R@P$_{0.9}$ & 0.4576 & 0.4861 & 6.23\% & 0.4893 & 0.5075 & 3.72\% \\ 
    \midrule
    \multirow{2}{*}{\makecell[l]{Shopee-large}} &
    AUC & 0.8785 & 0.8853 & 0.77\% & 0.8954 & 0.9091 & 1.53\% \\ &
    R@P$_{0.9}$ & 0.5245 & 0.5428 & 3.49\% & 0.5517 & 0.5783 & 4.82\% \\ 
    \midrule
    \multirow{2}{*}{\makecell[l]{Amazon}} &
    AUC & 0.8823 & 0.8951 & 1.45\% & 0.8975 & 0.9102 & 1.42\% \\ &
    R@P$_{0.9}$ & 0.2453 & 0.2566 & 4.61\% & 0.2767 & 0.2956 & 6.83\% \\ 
    \bottomrule
    \end{tabular}%
  \label{table: more_results}
\end{table}%
\vspace{-2mm}

\section{Visualization of {\mysystem}}
\label{sec: visualization}
\begin{figure}
\centering
\includegraphics[scale=0.27]{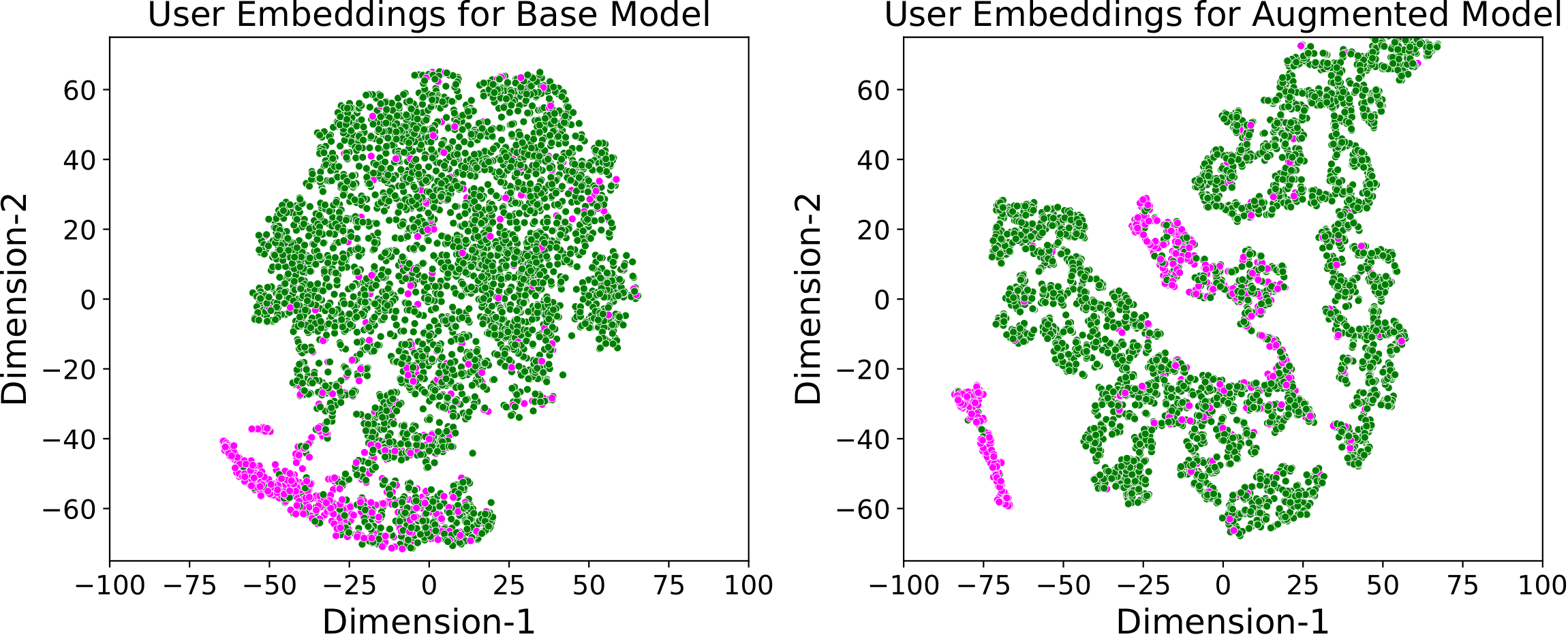}
\caption{Visualization of the first layer outputs of the output layer in the base BiLSTM model and its augmented model on the Amazon dataset. Fraudulent users are depicted in red, while normal users are depicted in green.}
\label{fig: visualization}
\end{figure}

We use t-SNE to visualize the user embeddings of a base model and its augmented version by {\mysystem}, as shown in Figure~\ref{fig: visualization}.
We apply the same t-SNE settings for both cases, with a Euclidean distance metric and a perplexity of 30. 
Unlike most existing methods that use contrastive loss on the encoder outputs, our proposed {\mysystem} applies contrastive loss on the predicted logits. As a result, the cohort augmentation has a stronger effect on the first layer output of the output layer, which we use for visualization.
From Figure~\ref{fig: visualization}, we observe that fraudulent users do not always form a single cluster, but rather are dispersed across multiple groups of varying sizes.
This shows the variety and complexity of fraud schemes and user behaviors.
Normal users also have different types of clusters, indicating their different preferences and habits.
We also notice that the augmented model forms more distinct and compact clusters than the original base model.
This is because the personalized cohort augmentation can capture the similarities and differences among users, and create more homogeneous and isolated small clusters.
Furthermore, the augmented model can better distinguish fraudulent cohorts from normal user groups, owing to both the enhanced clustering precision conferred by cohort augmentation and negative neighbor separation techniques.
These techniques can push away similar but negative neighbors during the cohort-augmented training process, and make each cluster more cohesive and discriminative.

\section{Hyper-parameters Settings on {\mysystem}}
\label{subsec: hyper_settings}
The hyperparameters $\alpha$ and $\beta$ play a crucial role in {\mysystem}'s objective function.
Specifically, they correspond to the weights of the supervised contrastive loss and auxiliary reconstruction loss, respectively. 
To provide a comprehensive understanding of how different settings of these hyperparameters affect the performance of {\mysystem}, we conducted a detailed analysis.
This analysis focused on evaluating the changes in model performance when varying the values of $\alpha$ and $\beta$.
The experiments utilize base models HEN~\cite{Zhu2020} and MINT~\cite{Xiao2023} with Shopee-small dataset.
The results are summarized in Table~\ref{table: alpha} and Table~\ref{table: beta}, and the analysis is presented as follows:

\begin{table}[t]
  \centering
  \caption{{\mysystem} performance across $\alpha$ values on the Shopee-small dataset.}
  \setlength{\tabcolsep}{1.3pt}
    \begin{tabular}{llccccccc }
    \toprule
    \multirow{2}{*}{\textbf{Model}} & 
    \multirow{2}{*}{\textbf{}} & 
    \multicolumn{7}{c}{\textbf{$\alpha$}} \\
    & & 0 & 0.00001 & 0.0001 & 0.001 & 0.01 & 0.1 & 1 \\
    \midrule
    \multirow{2}{*}{\makecell[l]{HEN}} &
    w/o & 0.8611 & 0.8611 & 0.8611 & \textbf{0.8611} & 0.8611 & 0.8611 & 0.8611 \\ &
    w/ & 0.8671 & 0.8693 & 0.8716 & \textbf{0.8730} & 0.8696 & 0.8654 & 0.8592 \\
    \midrule
    \multirow{2}{*}{\makecell[l]{MINT}} &
    w/o & 0.8892 & 0.8892 & 0.8892 & \textbf{0.8892} & 0.8892 & 0.8892 & 0.8892 \\ &
    w/ & 0.8941 & 0.8953 & 0.8962 & \textbf{0.8971} & 0.8954 & 0.8937 & 0.8895 \\
    \bottomrule
    \end{tabular}%
   \label{table: alpha}
\end{table}%

\begin{table}[t]
  \centering
  \caption{{\mysystem} performance across $\beta$ values on the Shopee-small dataset.}
  \setlength{\tabcolsep}{1.pt}
    \begin{tabular}{llccccccc }
    \toprule
    \multirow{2}{*}{\textbf{Model}} & 
    \multirow{2}{*}{\textbf{}} & 
    \multicolumn{7}{c}{\textbf{$\beta$}} \\
    & & 0 & 0.0000001 & 0.000001 & 0.00001 & 0.0001 & 0.001 & 0.01 \\
    \midrule
    \multirow{2}{*}{\makecell[l]{HEN}} &
    w/o & 0.8611 & 0.8611 & 0.8611 & \textbf{0.8611} & 0.8611 & 0.8611 & 0.8611 \\ &
    w/ & 0.8713 & 0.8722 & 0.8727 & \textbf{0.8730} & 0.8719 & 0.8678 & 0.8635 \\
    \midrule
    \multirow{2}{*}{\makecell[l]{MINT}} &
    w/o & 0.8892 & 0.8892 & 0.8892 & \textbf{0.8892} & 0.8892 & 0.8892 & 0.8892 \\ &
    w/ & 0.8958 & 0.8964 & 0.8969 & \textbf{0.8971} & 0.8953 & 0.8936 & 0.8913 \\
    \bottomrule
    \end{tabular}%
   \label{table: beta}
\end{table}%

\begin{itemize}[leftmargin=*]
\item
\textbf{Study of $\alpha$}: An increase in $\alpha$ from its optimal value of 0.001 to 0.01 results in a marginal performance decline for both HEN with {\mysystem} and MINT with {\mysystem}. 
A further increase of $\alpha$ to 1.0 leads to a more pronounced drop, due to an overemphasis on distinguishing training samples from negative neighbors, which could cause overfitting. 
On the other hand, a decrease in $\alpha$ below 0.001 reduces the model’s ability to counteract noise from negative neighbors, which will negatively impact {\mysystem}’s effectiveness.
\item
\textbf{Study of $\beta$}: Adjusting the hyperparameter $\beta$ also alters {\mysystem}’s performance.
A higher $\beta$ diminishes the effectiveness of {\mysystem}’s cohort augmentation by aligning the main encoder too closely with the auxiliary encoder.
This will limit the diversity and robustness that cohort augmentation is intended to introduce.
Conversely, a reduction in $\beta$ leads to a slight decrease in effectiveness, which is mitigated by the use of burn-in vectors from augmentation neighbors to enhance target sample representations.
These burn-in vectors help to enhance the representations of the target samples by introducing additional contextual information, which compensates to some extent for the decreased influence of the auxiliary reconstruction loss.
\end{itemize}

Overall, the performance of {\mysystem} remains relatively stable across different hyperparameter values. An $\alpha$ value of 0.001 and $\beta$ 0.00001 achieve the best performance across different models. Hence, the objective function can be easily optimized without laborious hyperparameter tuning.

\section{Computational Complexity Analysis of {\mysystem}}
\label{subsec: complexity}
The computational complexity of {\mysystem} involves both time and space complexity.
We denote the number of parameters $P$, the number of stored sample embedding vectors as $n$, the dimensionality of the embeddings as $d$, and the number of neighbors as $k$.
\begin{itemize}[leftmargin=*]
\item
\textbf{Time Complexity}: The vector burn-in step’s time complexity is determined by the base model, with a complexity of $O(T)$.
The cohort-augmented learning step’s time complexity is mainly dominated by the augmented learning model and the attentive neighbor aggregation layer, with a complexity of $O(T)$ and $O(k^2*d)$, respectively.
It also includes the search time for $k$ augmentation neighbors using the HNSW index, with a complexity of $O(k*log(n))$.
So the time complexity should be $O(T + k^2*d + k*log(n))$.
\item
\textbf{Space Complexity}: The space complexity in the vector burn-in step is $O(P)$.
The cohort-augmented learning step involves storing burn-in vectors and constructing a graph-based index, leading to a complexity of $O(nlog(n) + nd + P)$. 
Therefore, the overall space complexity is $O(nlog(n) + nd + P)$.
\end{itemize}

Given that $k$ is set to 5, the embedding dimension $d$ is 64, and $log(n)$ is usually small, the time complexity is mainly determined by the model rather than cohort neighbor searching.
The space complexity is dominated by the stored vector size and the model size, which are typically moderate in our anti-fraud system.
\clearpage

\end{document}